\documentclass[9pt,twocolumn,twoside]{pnas-new}

\usepackage{graphicx}			
\usepackage{subfigure}
\usepackage{float}
\usepackage{amsmath,amssymb,bm}
\usepackage{mathtools}

\templatetype{pnasresearcharticle} 

\title{Skin cancer detection based on deep learning and entropy to detect outlier samples}

\author[a,b,1]{Andre G. C. Pacheco}
\author[b,c,1]{Abder-Rahman Ali}
\author[b,1]{Thomas Trappenberg}

\affil[a]{Graduate Program in Computer Science, Federal University of Espírito Santo, Brazil}

\affil[b]{Faculty of Computer Science, Dalhousie University,
Canada}

\affil[c]{Faculty of Natural Sciences, Computing Science and Mathematics, University of Stirling, United Kingdom}

\dates{Manuscript presented to the ISIC challenge @ MICCAI2019 Workshop on August 23rd, 2019}

\doi{ISIC challenge @ MICCAI2019}

\leadauthor{Pacheco, Ali and Trappenberg} 

\correspondingauthor{\textsuperscript{1}E-mails: agcpacheco@inf.ufes.br, abder@cs.stir.ac.uk and tt@dal.cs.ca}


\begin{abstract}
We describe our methods that achieved the 3rd and 4th places in tasks 1 and 2, respectively, at ISIC challenge 2019. The goal of this challenge is to provide the diagnostic for skin cancer using images and meta-data. There are nine classes in the dataset, nonetheless, one of them is an outlier and is not present on it. To tackle the challenge, we apply an ensemble of classifiers, which has 13 convolutional neural networks (CNN), we develop two approaches to handle the outlier class and we propose a straightforward method to use the meta-data along with the images. Throughout this report, we detail each methodology and parameters to make it easy to replicate our work. The results obtained are in accordance with the previous challenges and the approaches to detect the outlier class and to address the meta-data seem to be work properly.
\end{abstract}

\begin{document}

\maketitle
\thispagestyle{firststyle}
\ifthenelse{\boolean{shortarticle}}{\ifthenelse{\boolean{singlecolumn}}{\abscontentformatted}{\abscontent}}{}

\section{Introduction}
Skin cancer incidences have been increasing throughout the last decade \cite{ACS2019}. Unfortunately, individual cases of cancer are not required to be reported by most cancer registries \cite{CancerStats2019}. However, the World Health Organization (WHO) estimates that around 3 million skin cancers occur globally each year \cite{WHO2019}. 

The use of computer-aided diagnosis (CAD) systems for skin cancer detection has been increasing over the past decade. Recently, deep learning models have been achieving remarkable results in different medical image analysis tasks \cite{litjens2017}. In particular, convolutional neural networks (CNN) have become the standard approach to handle this kind of problem \cite{tajbakhsh2016}. The progress is largely due to the International Skin Imaging Collaboration (ISIC) \cite{codella2019skin}, which provides a large skin cancer dataset to the research community.

In this report, we present our strategies for the ISIC challenge 2019. We describe the models used, the major difficulty we encountered with the tasks and the results that we achieved. The rest of this manuscript is organized as follows: first we describe the dataset and the tasks characteristics; next we present the methods adopted to tackle both tasks; Lastly we show the achieved results.

\section{ISIC 2019}
\subsection{Dataset} 
\begin{figure}
\centering    
\subfigure{
\includegraphics[width=0.45\columnwidth]{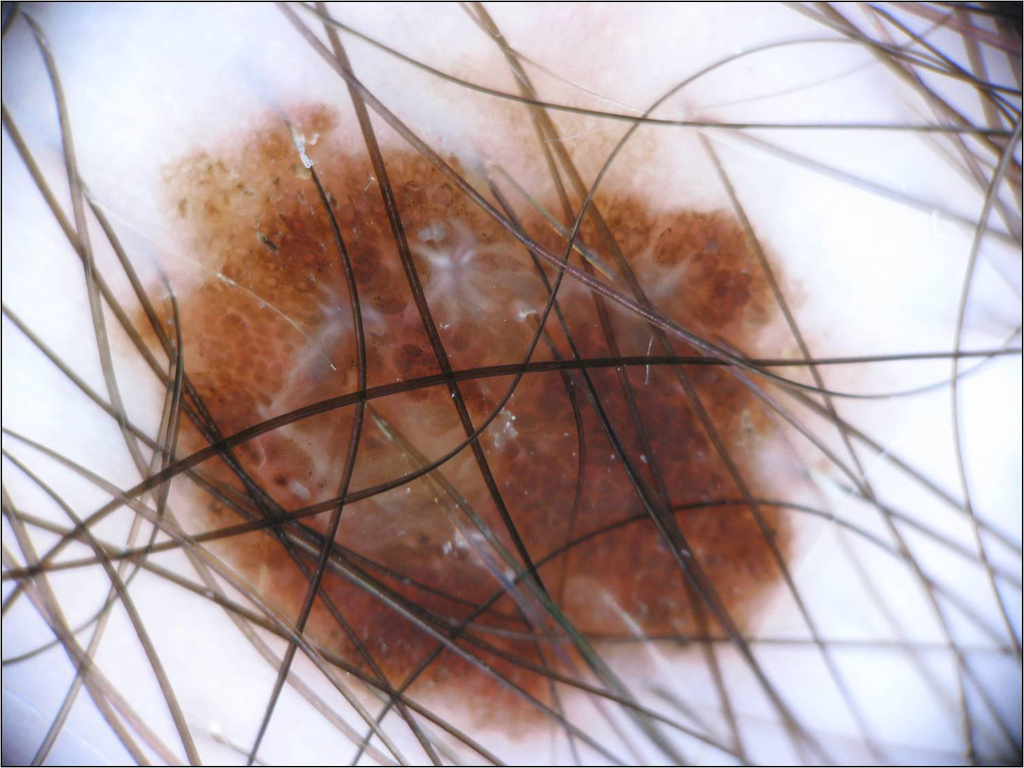}
}
\quad
\subfigure{
\includegraphics[width=0.45\columnwidth]{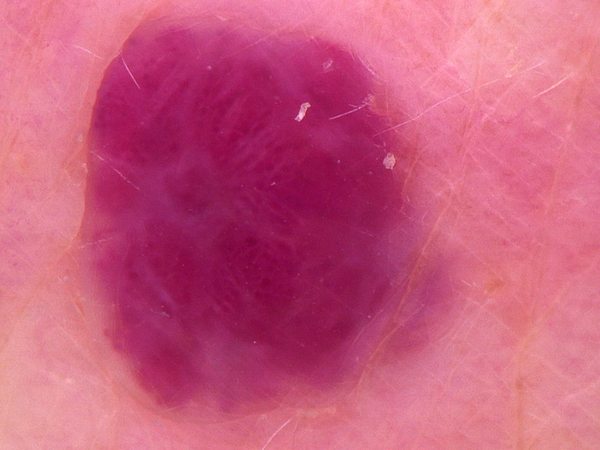}
}
\subfigure{
\includegraphics[width=0.45\columnwidth]{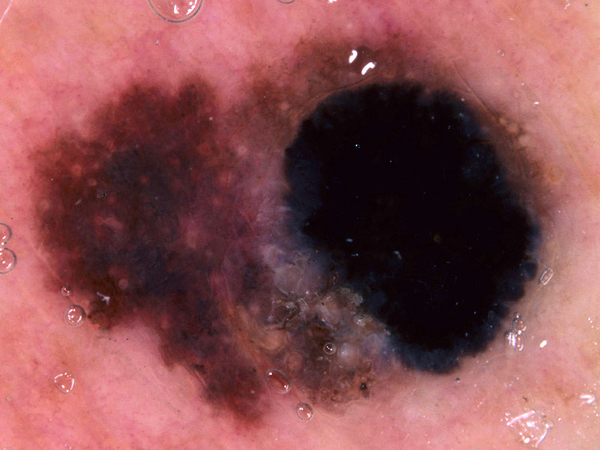}
}
\quad
\subfigure{
\includegraphics[width=0.45\columnwidth]{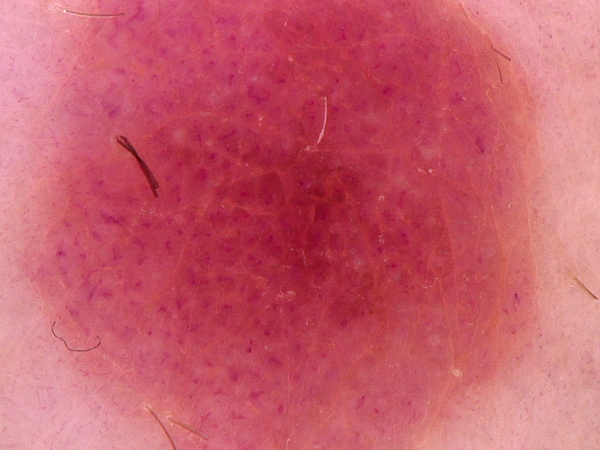}
}
\caption{Samples of skin diseases from the ISIC 2019 dataset}
\label{fig:train_samples}
\end{figure}

With the aim of both supporting clinical training and further technical research, which will eventually lead to automated algorithmic analysis, the International Skin Imaging Collaboration (ISIC) developed an international repository of dermoscopic images known as the ISIC Archive\footnote{https://www.isic-archive.com}. Every year the ISIC increases its archive and promote a challenge to leverage the automated skin cancer detection. For ISIC 2019, 25,331 dermoscopy images are available for training across 8 different categories. The test dataset is composed of 8,239 images and contains an additional outlier class not represented in the training data, which the new systems must be able to identify. Beside the images, the dataset also contains meta-data for most of the images. The meta-data is composed of the patient's age and sex, and the region where the individual with the skin lesion is located. All these data come from the BCN\_20000 (Department of Dermatology, Hospital Clínic de Barcelona) \cite{Combaliaa2019}, HAM10000 \cite{TschandlS2018} (ViDIR Group, Department of Dermatology, Medical University of Vienna), and from an Anonymous resources \cite{Codellaa2017}. In Fig. \ref{fig:train_samples} is shown some samples of skin diseases from the ISIC 2019 dataset.

\subsection{Tasks description}
The ultimate goal of both tasks is to provide the diagnostic for the dermoscopy images among nine different diagnostic categories: melanoma (MEL), melanocytic nevus (NV), basal cell carcinoma (BCC), actinic keratosis (AK), benign keratosis (BKL), dermatofibroma (DF), vascular lesion (VASC), squamous cell carcinoma (SCC), and none of the others (UNK). The UNK class is an outlier distribution that is not present in the trained dataset. The number of samples for each class in the training dataset is described in Table \ref{tab:freq_labels}. The difference between both task is related to the meta-data. While for task 1 it is not allowed to use this information, for task 2 it is required.

\begin{table}
\centering
\caption{The number of samples for each class in the training dataset}
\begin{tabular}{cc}
\hline
\textbf{Diagnostic}                 & \textbf{Number of samples} \\ \hline
\multicolumn{1}{c|}{\textit{MEL}}   & 4522                       \\
\multicolumn{1}{c|}{\textit{NV}}    & 12875                      \\
\multicolumn{1}{c|}{\textit{BCC}}   & 3323                       \\
\multicolumn{1}{c|}{\textit{AK}}    & 867                        \\
\multicolumn{1}{c|}{\textit{BKL}}   & 2624                       \\
\multicolumn{1}{c|}{\textit{DF}}    & 239                        \\
\multicolumn{1}{c|}{\textit{VASC}}  & 253                        \\
\multicolumn{1}{c|}{\textit{SCC}}   & 628                        \\
\multicolumn{1}{c|}{\textit{UNK}}   & 0                          \\ \hline
\multicolumn{1}{c|}{\textbf{Total}} & \textbf{25,331}            \\ \hline
\end{tabular}
\label{tab:freq_labels}
\end{table}

\section{Methods}
In this section, we describe our strategy to address both tasks. First we describe our approach to classify the 8 skin lesions. Next, we present our methods to detect the outlier class. Lastly, we describe how we used the meta-data.

\subsection{The skin cancer classification} \label{sec:skin_detection}
We start to address task 1 by classifying the eight known classes. It is known from the previous ISIC challenges that the most successful approaches are those based on ensemble of classifiers \cite{tschandl2019}. For this reason, we adopted the following convolutinal neural networks (CNNs): SENet \cite{hu2018squeeze}, PNASNet \cite{liu2018progressive}, InceptionV4 \cite{szegedy2017inception}, ResNet-50/101/152 \cite{he2016deep}, DenseNet-121/169/201 \cite{huang2017densely}, MobileNetV2 \cite{sandler2018mobilenetv2}, GoogleNet \cite{szegedy2015}, and VGG-16/19 \cite{simonyan2014}.

In order to train all networks, the convolutional layers were kept and only the classifiers were changed to fit the task requirements. In addition, all models were pretrained on ImageNet \cite{russakovsky2015imagenet}. All models were fine-tuned for 150 epochs using the Adam optimization \cite{kingma2014adam} with a starting learning rate equal to $0.0001$ and the batch size equal to 32. The learning rate is scheduled to be reduced by a factor of 0.2 if the models fail to improve the validation loss for 10 epochs. Finally, we use early stopping, also based on a stagnant validation loss for 15 epochs.

As can be seen in Table \ref{fig:train_samples}, the dataset is very imbalanced. To address this issue, we used a weighted version of the cross-entropy as the loss function. The classes were weighted according to their frequency, i.e., the more the number of samples the lower the weight. We also tried to use upsampling to equalize the number of samples for each class, but it created a high bias for most classes, which resulted in a lower performance compared to the weight loss function approach.

All images were resized to $229 \times 229$ to InceptionV4, $331 \times 331$ to PNASnet, and $224 \times 224$ for the remaining networks. In addition, we applied data augmentation using common image processing operations. We adjust brightness, contrast, saturation and hue, and we apply horizontal and vertical rotations, translations, re-scale and shear. Also, before the data augmentation, we applied the shades of gray method \citep{finlayson2004shades} for all images.

\subsubsection{The ensemble of CNNs}
We consider two ensembles for this task. The first one is composed of the 13 models presented in the previous subsection. The second one consists of the best three models according to the balanced accuracy. In order to aggregate the model, we considered the following approach: majority voting, maximum probability, and average of the probabilities. The approach that worked best for us was the last one, the average of the probabilities. The results for each model and ensembles will be presented in section \ref{sec:results}.

\subsection{Handling the outlier class}
The main part of this task is to detect the outlier class. To deal with it, we propose two approaches: a hierarchical classifier and a outlier selection based on entropy estimation. We describe each of them in the following.

\subsubsection{Approach 1: hierarchical classifier}
To handle with the outlier class, our first approach adopted is a hierarchical classifier. This approach assumes that the outlier class contains only skin images\footnote{This piece of information was provided by the ISIC organization team through the ISIC forum.}. Thus, the classifier is used to differentiate skin images from lesions. While this requires some form of knowledge of the outlier class, we can also treat this approach as a kind of base-line to compare other approaches. 

\begin{figure}
\centering    
\subfigure[Original image]{
\includegraphics[width=0.45\columnwidth]{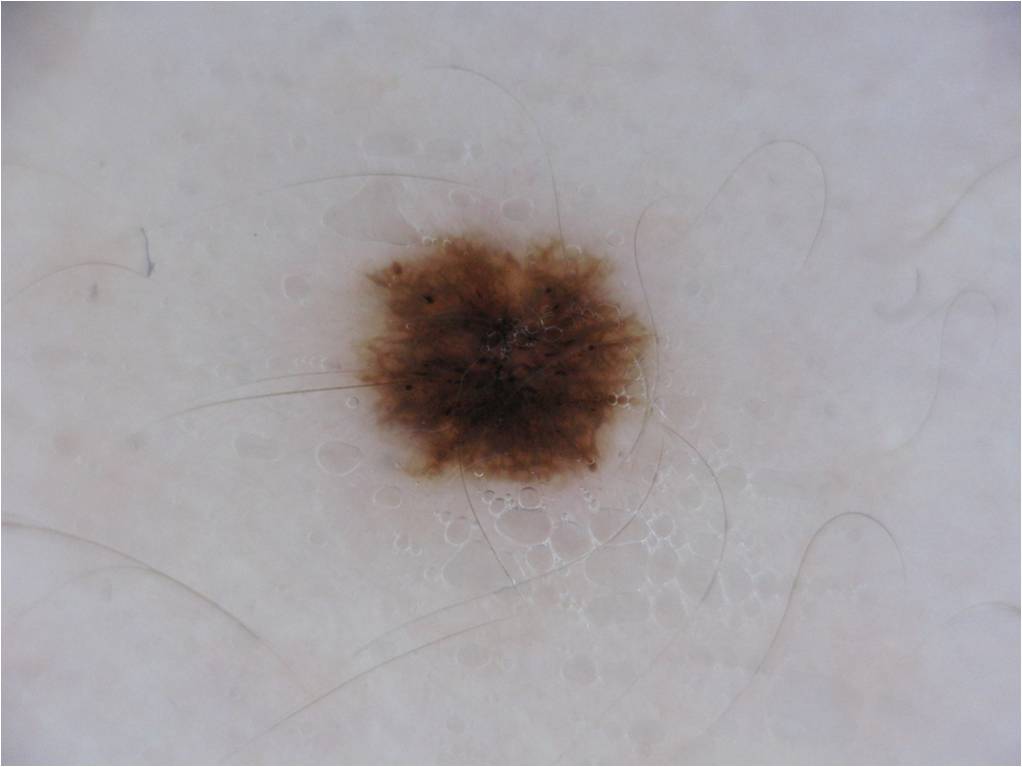}
}
\quad
\subfigure[Patched image]{
\includegraphics[width=0.45\columnwidth]{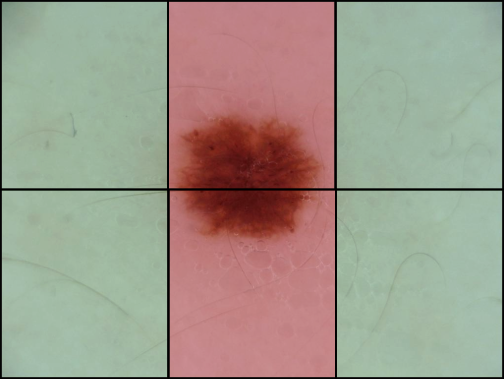}
}
\caption{An illustration of our method to obtain skin images from the original images}
\label{fig:patches_for_hierarchical}
\end{figure}

In order to obtain skin samples, we developed a script to split the images in different patches. Next, we select the patches that contain only skin. This method is illustrated in Fig. \ref{fig:patches_for_hierarchical}, in which the green patches are selected and the red ones are rejected. After performing this script, we selected 1,239 patches of skin to train the classifier. There is no external data included in this approach.

We adopted a ResNet-50 architecture as the classifier for this approach. Its training phase is carried out as described in section \ref{sec:skin_detection}. This is an easy task for a CNN. The model achieve an accuracy of 99\% in the test partition. However, we may point out two weakness: 1) we do not have a large variety of skin images, which may bias the final result; 2) it works only for skins and is unable to detect any extra outlier. For this reason, we use it as a baseline for the other ones.

\subsubsection{Approach 2: entropy selection}
In order to detect any type of outlier, we propose an approach based on the Shannon Entropy \cite{shannon1948mathematical} to detect the unknown class. The classifiers used in this work output the probabilities for each class using the softmax function. Basically, when a classifier is in doubt about a new sample, and it may be an outlier, it assigns a value for the probability of different classes. On the other hand, when the classifier is certain about its decision, it assigns a high probability for only one class. As a result, the entropy for the first case is higher than that one for the last case. This approach exploits this point. 

Considering an array $\mathbf{x}$ of probabilities provided by the output of a classifier, the entropy is computed as follows:

\begin{equation}
    h(\mathbf{x}) = - \sum \mathbf{x} \log_2 \mathbf{x}
\end{equation}

\noindent First, we compute the average and the maximum entropy, $\bar{h}_{hit}$ and $\breve{h}_{hit}$, respectively, for those samples that are corrected classified. Next, we do the same for those samples that are miss classified, which produces $\bar{h}_{miss}$ and $\breve{h}_{miss}$. Every class will have its own $\bar{h}_{hit}$, $\breve{h}_{hit}$, $\bar{h}_{miss}$ and $\breve{h}_{miss}$ values. It means we compute the entropies values locally instead of globally. In addition, we select these values in the validation set to be applied in the test set. The step-by-step to identify an outlier sample in the test dataset using this approach is described as follows:

\begin{enumerate}
    \item Using the validation set, compute $\bar{h}_{hit}$, $\breve{h}_{hit}$, $\bar{h}_{miss}$ and $\breve{h}_{miss}$ for each class in the dataset.
    \item Compute the entropy $h_s$ for all samples in the test dataset.
    \item Based on the prediction test, if $h_s$ is greater than $\bar{h}_{hit}$ and $\bar{h}_{miss}$, this may be a outlier.
    \item If the entropy of the samples selected in the previous step greater than $\frac{\breve{h}_{hit} + \breve{h}_{miss}}{2}$, we consider this sample as unknown.
\end{enumerate}

\noindent Applying this step-by-step, many true lesions were being identifying as unknown. We realized that classes such as, SCC/BCC and NV/MEL, are very similar and, eventually, the classifier assigns a value for both probabilities. Thus, we realized that we need to take into account the relationship between those classes. In addition, beyond the entropy, we also compute the average probability for each class considering the hit ($\bar{p}_{hit})$ and miss ($\bar{p}_{miss})$ groups. Next, we add one more step in the algorithm:

\begin{enumerate}
    \item[5] Based on the $\bar{p}_{hit}$ and $\bar{p}_{miss}$ for the predicted class and considering $p_s$ the probabilities obtained by the current new sample, compute the cosine similarity between theses arrays as follow:
    \begin{equation}
        \textrm{sim}_\textrm{cos}(\mathbf{x}, \mathbf{y}) = \frac{\mathbf{x} \mathbf{y}}{\parallel \mathbf{x} \parallel \parallel \mathbf{y} \parallel }
    \end{equation}
    Finally, if $\textrm{sim}_\textrm{cos}(\bar{p}_{miss}, p_s)$ is greater than $\textrm{sim}_\textrm{cos}(\bar{p}_{hit}, p_s)$, this sample is considering an outlier, otherwise it is a lesion.
\end{enumerate}

\subsection{Working with meta-data}
As stated before, for task 2 we must use the meta-data in order to provide the skin cancer diagnosis. The meta-data is composed for three pieces of information: the patient's age, the region of the body in which the lesion occurs and the patient's sex. It is worth mentioning that not all images contain meta-data. For 2,581 images, at least one of the three pieces of information is missing.

\begin{figure*}
\centering    
\subfigure[Boxplots for the patients' age per diagnostic]{
\includegraphics[scale=0.4]{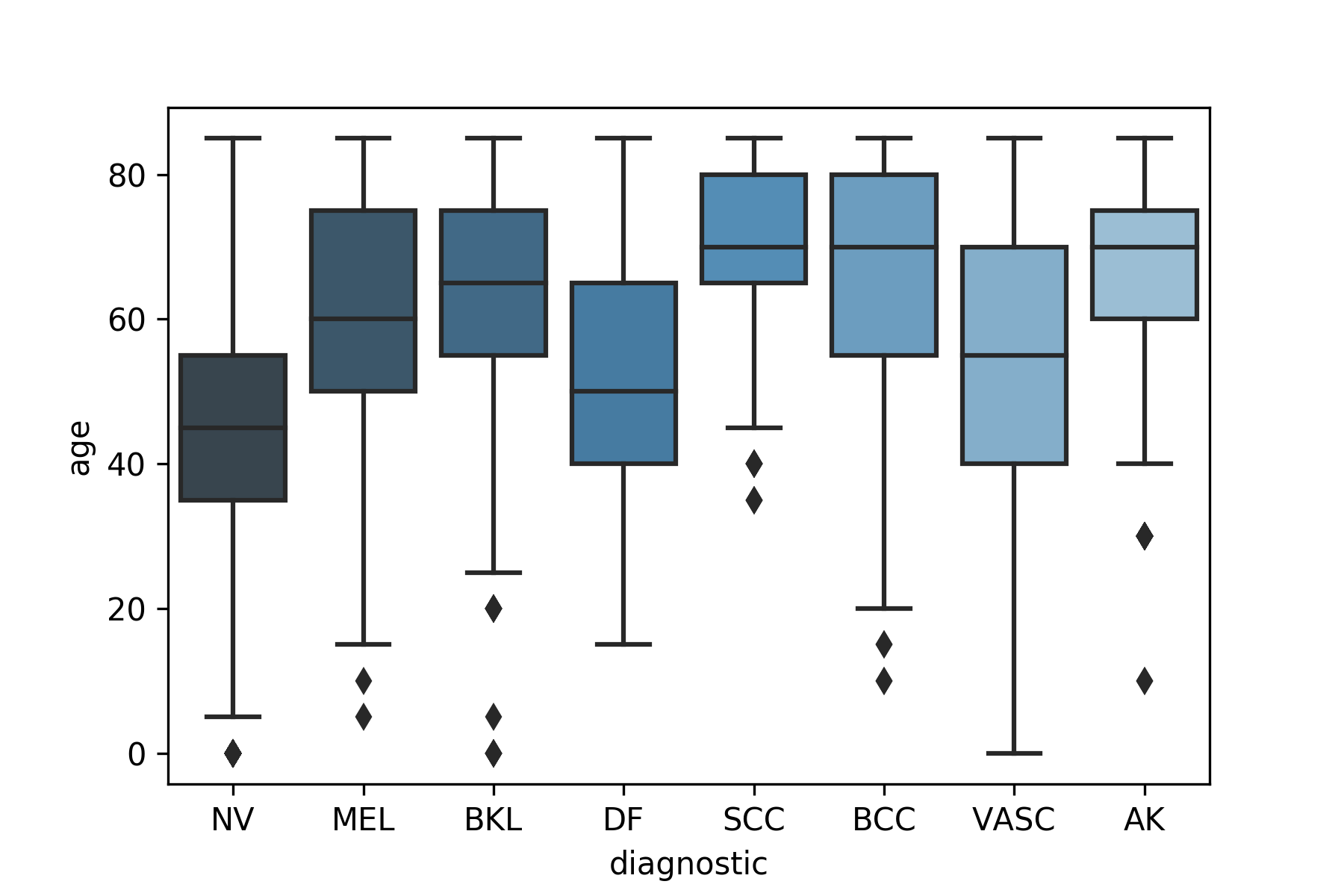}
}
\quad
\subfigure[The frequency of each sex per diagnostic]{
\includegraphics[scale=0.4]{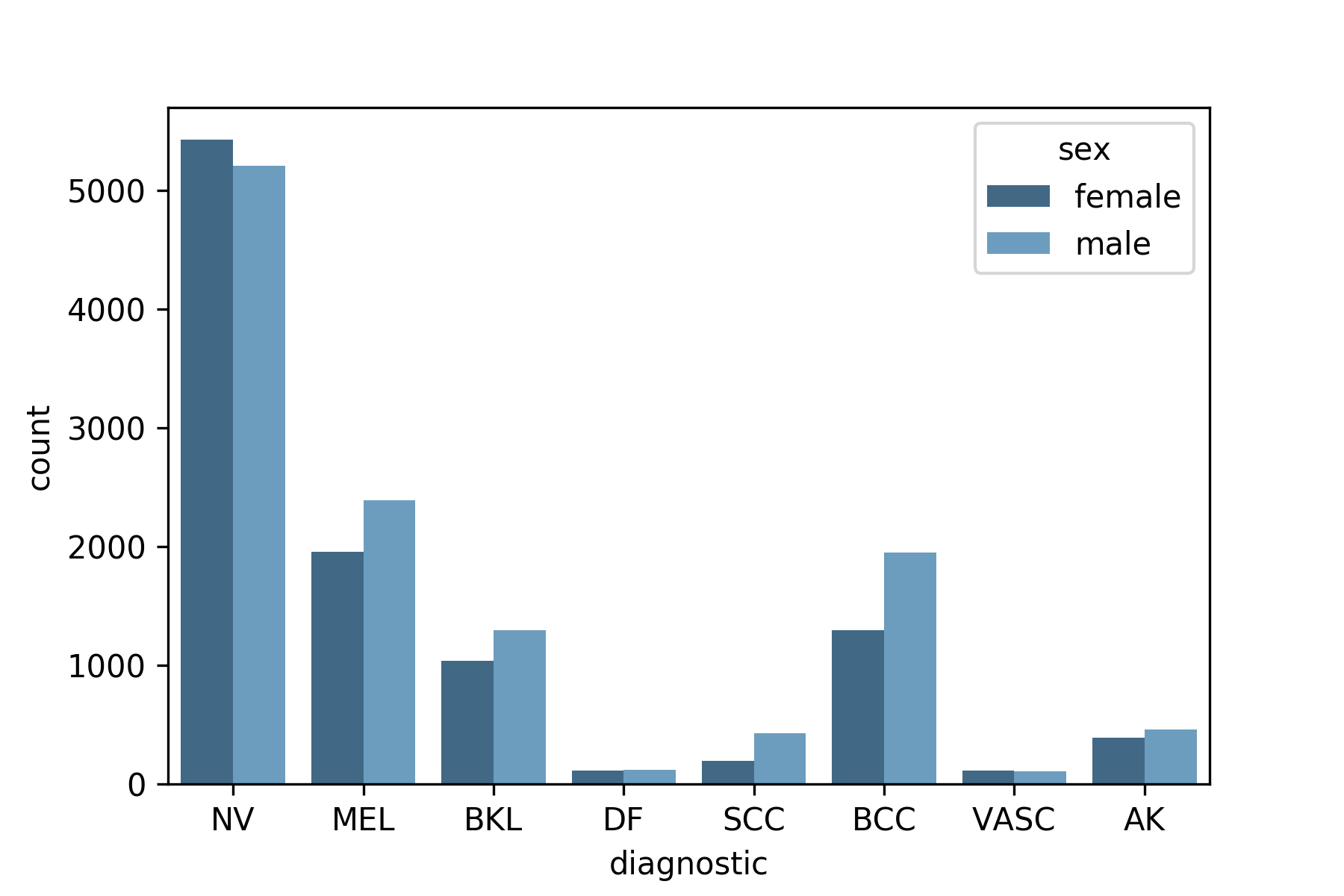}
}
\subfigure[The frequency of each region per diagnostic]{
\includegraphics[scale=0.55]{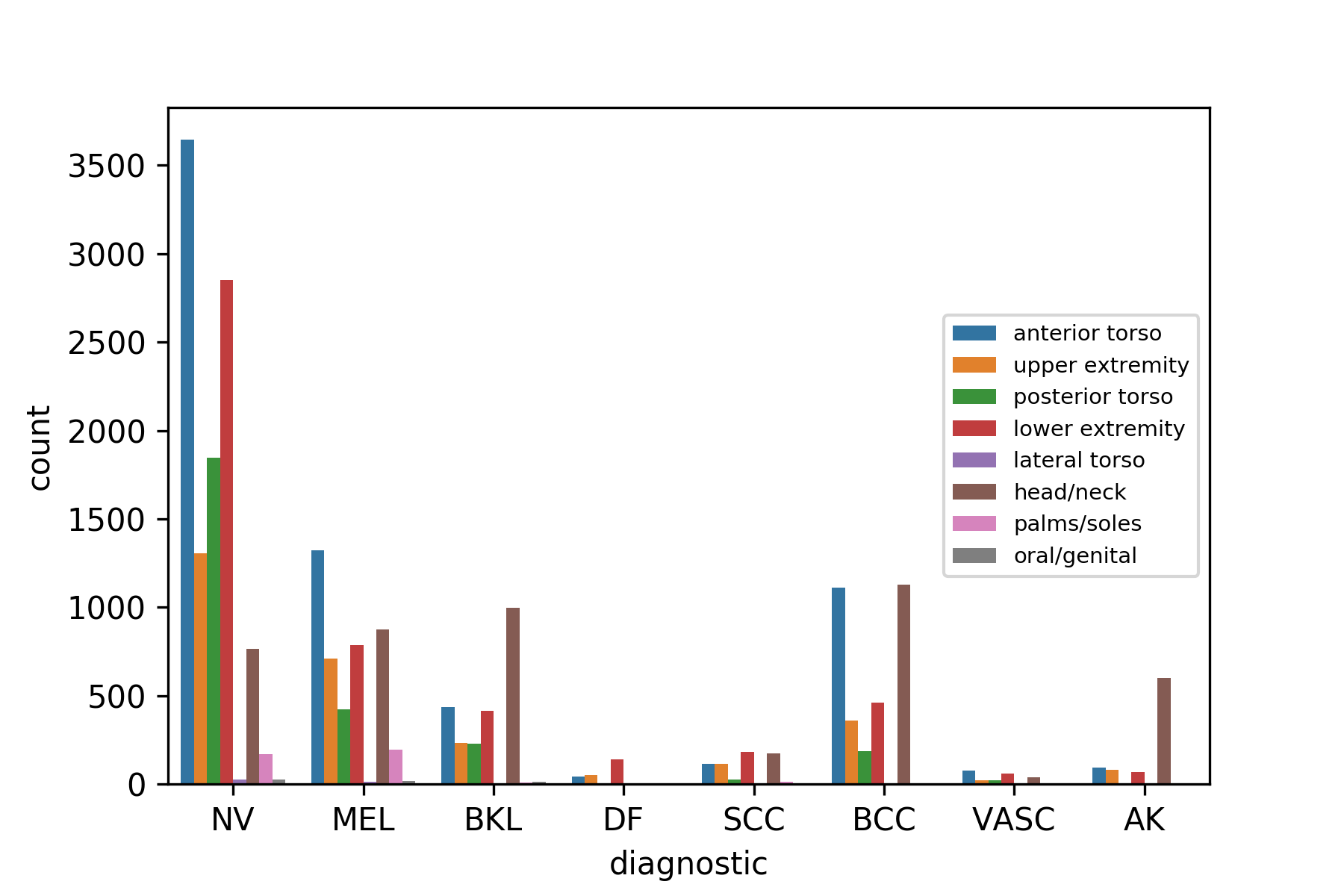}
}

\caption{Data exploration plots}
\label{fig:data_expo}
\end{figure*}

In order to better understand the meta-data characteristics, we performed a data exploration analysis for these pieces of information. In Fig. \ref{fig:data_expo} is depicted the plots for each type of meta-data. We may observe that the age may be helpful to differentiate NV, MEL and BKL; the frequency for both sex is almost the same; and regarding the region, MEL and NV share similar regions, and AK is more frequent in head/neck.

Based on the data exploration, we developed a straightforward approach to consider the meta-data in the classification. First we use the histogram method \cite{hayter2012probability} to estimate the following probabilities:
\begin{itemize}
    \item $p_{FA}$: $p(\textrm{class}|\textrm{age and sex=female})$
    \item $p_{MA}$: $p(\textrm{class}|\textrm{age and sex=male})$
    \item $p_{FR}$: $p(\textrm{class}|\textrm{region and sex=female})$
    \item $p_{MR}$: $p(\textrm{class}|\textrm{region and sex=male})$
\end{itemize}

\noindent In the following, we compute the average probability ($\bar{p}(\textrm{class})$) outputted by a given model for each class. Then, we perform the following steps:

\begin{enumerate}
    \item Given a sample, obtain the probabilities outputted by the model for the top 2 predicted classes, $p_1(\textrm{class})$ and $p_2(\textrm{class})$, respectively.
    
    \item If $p_1(\textrm{class}) < \bar{p}(\textrm{class})$, go to the step 3, otherwise, select a new sample and return to step 1.
    
    \item Based on the top 2 predicted classes and the sex, obtain $p_{FA}$ and  $p_{FR}$ if the patient is female or $p_{MA}$ and  $p_{MR}$ if the patient is male. Next, compute the average between both probabilities ($\bar{p}_{AR}$). Do it for predicted class 1 ($\bar{p}_{AR_1}$) and 2 ($\bar{p}_{AR_2}$).
    
    \item Finally, increase the top 2 classes probabilities as follows: $p_1(\textrm{class}) = p_1(\textrm{class}) + \bar{p}_{AR_1}$ and $p_2(\textrm{class}) = p_2(\textrm{class}) + \bar{p}_{AR_2}$. If the new $p_2(\textrm{class})$ is greater than the new $p_1(\textrm{class})$, the classification for this sample becomes the class that was the second option.
    
\end{enumerate}

\noindent To handle the missing data, all prior probabilities are acquired considering only the samples that have all meta-data available. During the evaluation, if the new sample has a missing data, we compute the probability only for the available data. For example, if the age is available but the region is missing, we carry out the algorithm considering the probabilities for region equal to zero. If there is no meta-data available for the sample, we skip this method and consider only the probabilities obtained by the CNN(s) model.

Beyond this method, we also tried to apply the Naive Bayes, a decision tree, and convert the categorical features using one hot encode in order to concatenate them in the CNN classifier. None of these approaches improved the results obtained without using meta-data. Thus, we decided to propose the described method.

\section{Results} \label{sec:results}
In this section, we present the results obtained for the task 1 and 2. To test the models, we split the dataset into 80\% for training, 10\% for validation and 10\% for testing. We select the models based on the epoch in which it achieved the best validation loss. First we present the results considering the eight known classes. Next, we show the outlier detection performance for both approaches, with the classifier that included skin training, and with the entropy approach. Finally, we include the meta-data in the ensembles.

\begin{figure*}
\centering    
\subfigure[Ensemble 2 + meta-data]{
\includegraphics[scale=0.51]{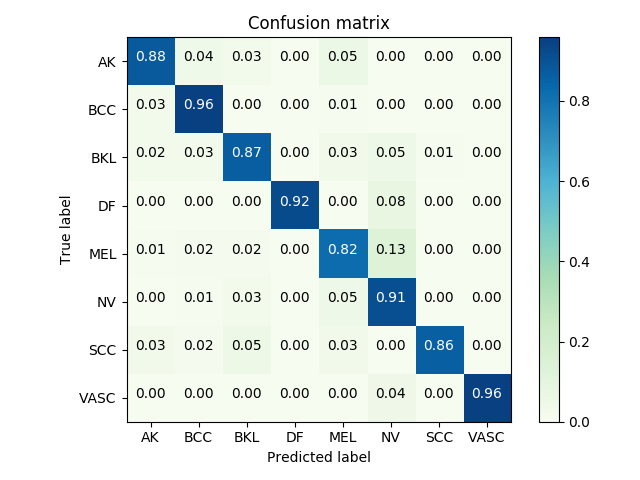}
}
\subfigure[Ensemble 2]{
\includegraphics[scale=0.51]{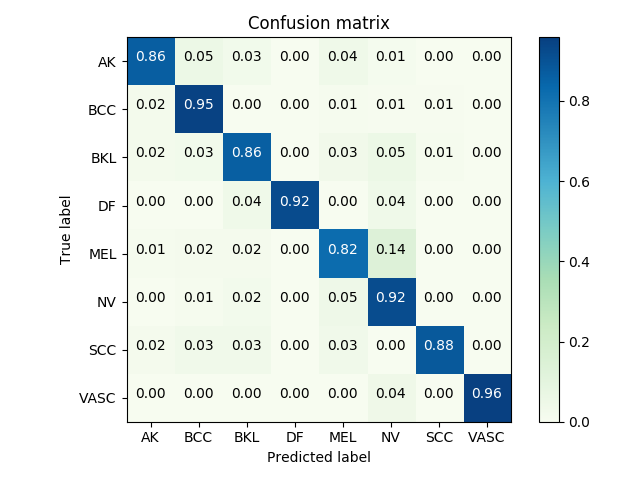}
}
\caption{The confusion matrices for ensemble 2 with and without using meta-data}
\label{fig:conf_mats}
\end{figure*}

In Table \ref{tab:results} is described the performance, in terms of balanced accuracy, for each model, for the ensembles, and for the ensembles + meta-data. We observe that the results using the meta-data is slightly better in terms of balanced accuracy. In addition, we can note, both ensembles present a balanced accuracy that is competitive with results achieved in the previous challenge. In Fig. \ref{fig:conf_mats} is depicted the confusion matrix for ensemble 2 with and without considering the meta-data. We may observe small differences in AK, BCC, BKL and NV detection.

\begin{table}
\centering
\caption{The results obtained for each model and for the ensembles. The ensemble 1 is composed of all models and the ensemble 2 consists of the best three models based on the balanced accuracy (in bold)}
\begin{tabular}{c|ccc}
\hline
\textbf{Model}        & \textbf{Balanced accuracy} & \textbf{Accuracy} & \textbf{AUC} \\ \hline
\textit{DenseNet-121} & 0.832                      & 0.840             & 0.974        \\
\textit{DenseNet-169} & 0.811                      & 0.830             & 0.96         \\
\textit{DenseNet-201} & 0.821                      & 0.851             & 0.975        \\
\textit{GoogleNet}    & 0.814                      & 0.820             & 0.966        \\
\textit{InceptionV4}  & 0.823                      & 0.831             & 0.971        \\
\textit{MobileNetV2}  & 0.812                      & 0.799             & 0.964        \\
\textit{PNASNet}      & \textbf{0.837}                      & 0.852             & 0.978        \\
\textit{ResNet-50}    & 0.820                      & 0.828             & 0.967        \\
\textit{ResNet-101}   & 0.812                      & 0.820             & 0.969        \\
\textit{ResNet-152}   & 0.818                      & 0.837             & 0.969        \\
\textit{SENet}        & \textbf{0.855}                      & 0.860             & 0.974        \\
\textit{VGG-16}       & 0.825                      & 0.807             & 0.968        \\
\textit{VGG-19}       & \textbf{0.842}                      & 0.827             & 0.972        \\ \hline
\textbf{Ensemble 1}   & 0.883                      & 0.890             & 0.988        \\
\textbf{Ensemble 2}   & 0.897                      & 0.910             & 0.989        \\ \hline

\textbf{Ensemble 1 + Meta-data}   & 0.891                      & 0.896             & 0.983        \\
\textbf{Ensemble 2 + Meta-data}   & 0.901                      & 0.910             & 0.987        \\ \hline

\end{tabular}
\label{tab:results}
\end{table}

\begin{figure}
\centering    
\subfigure{
\includegraphics[width=0.4\columnwidth]{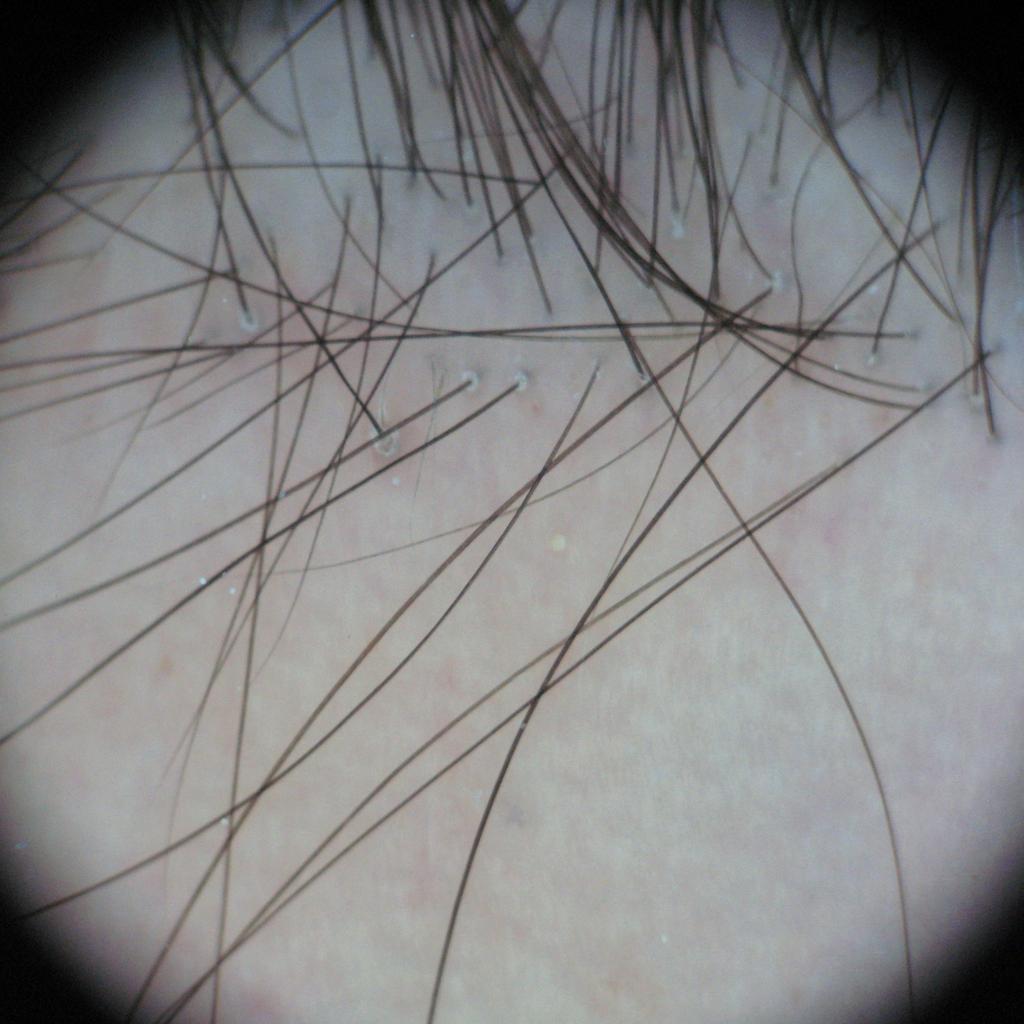}
}
\quad
\subfigure{
\includegraphics[width=0.4\columnwidth]{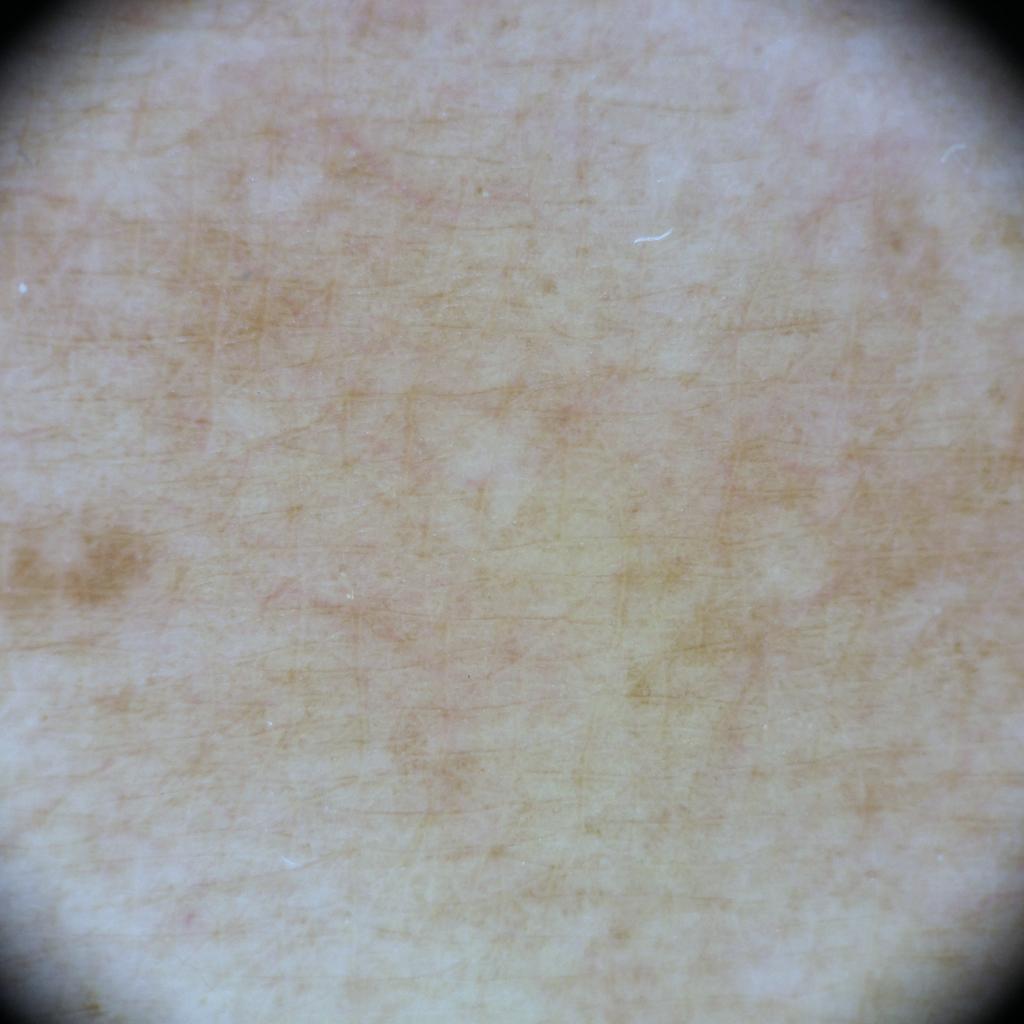}
}
\subfigure{
\includegraphics[width=0.4\columnwidth]{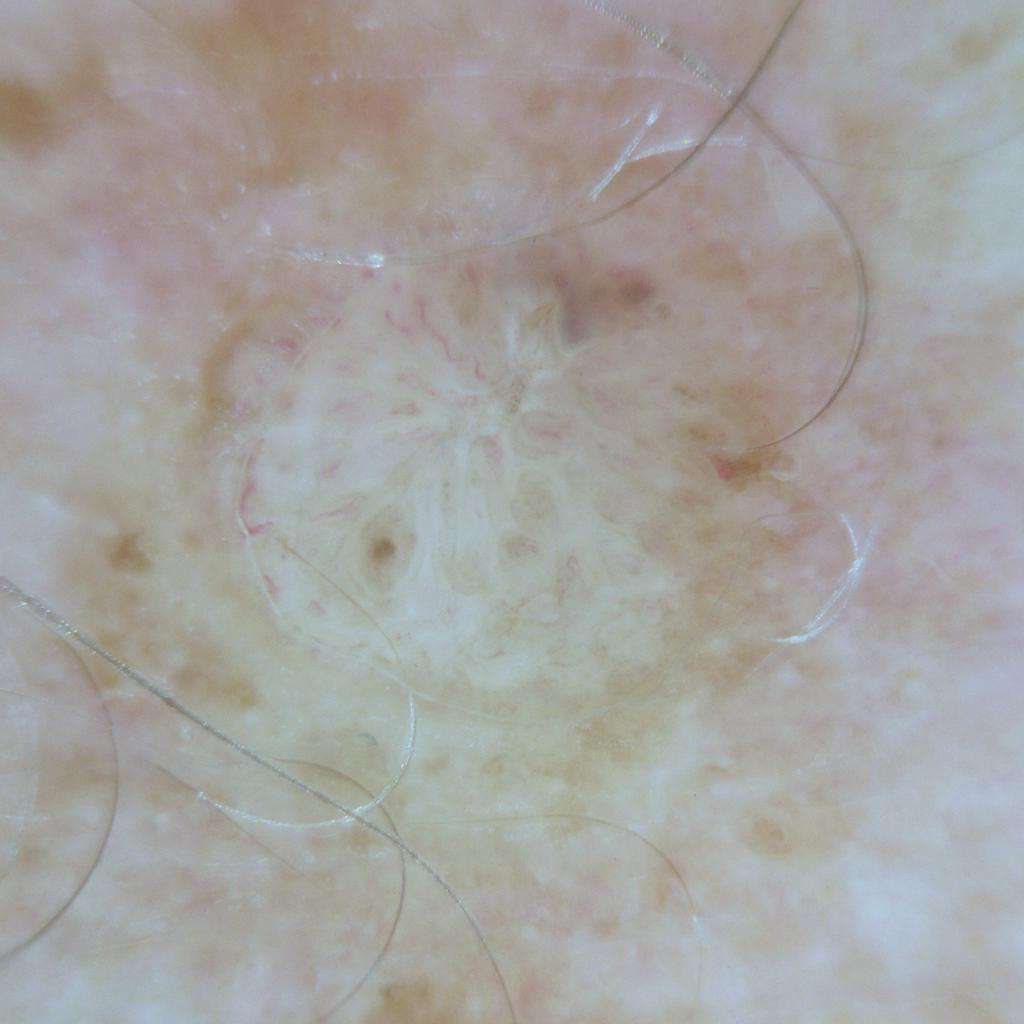}
}
\quad
\subfigure{
\includegraphics[width=0.4\columnwidth]{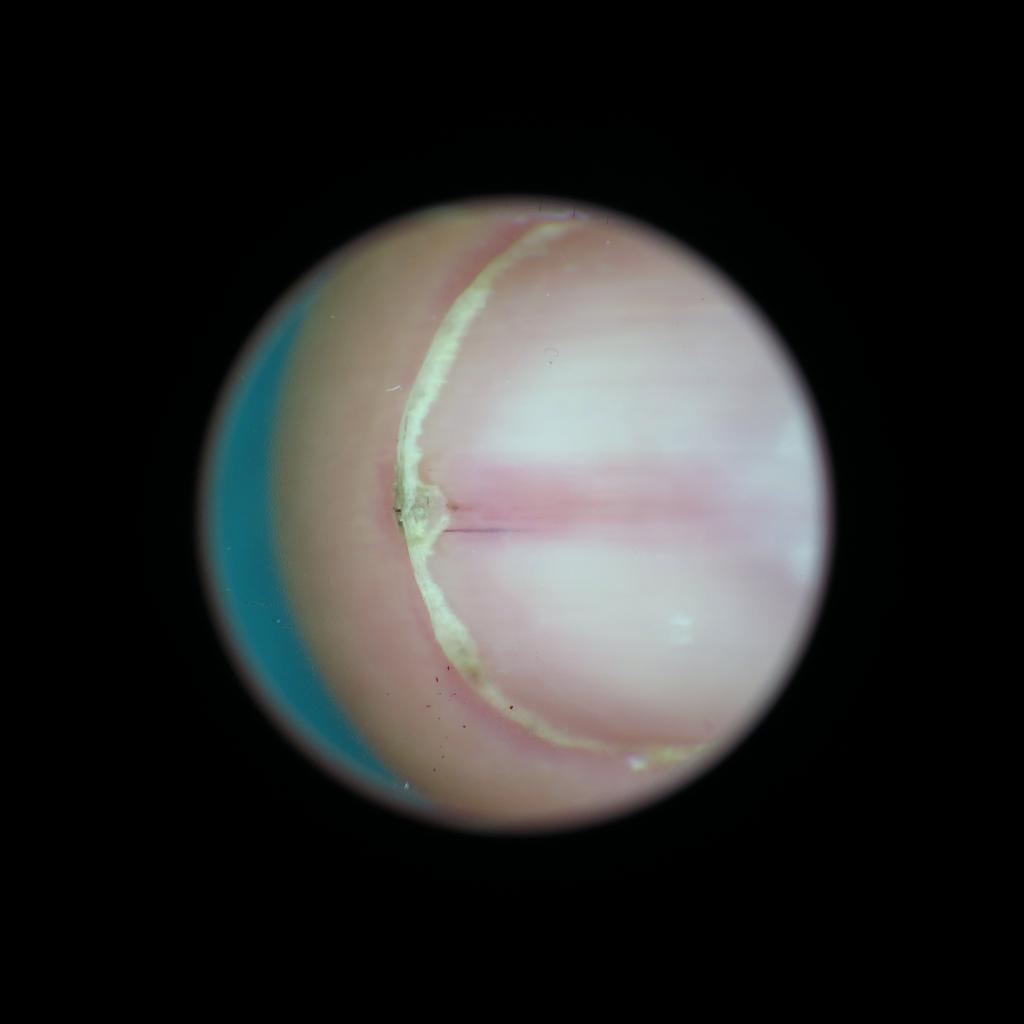}
}
\caption{Examples of images detected as outliers}
\label{fig:outiers}
\end{figure}

Regarding the outlier detection, we generated both ensembles considering the 8,238 final test images. In Table \ref{tab:outliers} is described the number of outliers that was found by each approach for each ensemble. In Fig. \ref{fig:outiers} is depicted some samples of outliers that were found by both approaches. Although we cannot ensure it is not a skin disease, the examples depicted seem plausible.

\begin{table}
\centering
\caption{The number of outliers found by each approach for each ensemble}
\begin{tabular}{cccc}
\hline
\textbf{Approach}                 & \textbf{Ensemble} & \textbf{Number of outliers} & \textbf{\%} \\ \hline
\multicolumn{1}{c|}{Hierarchical} & 1 and 2               & 45                          & 0.54        \\
\multicolumn{1}{c|}{Entropy}      & 1                 & 944                         & 11.45       \\
\multicolumn{1}{c|}{Entropy}      & 2                 & 579                         & 7.02        \\ \hline
\end{tabular}
\label{tab:outliers}
\end{table}

\section{Final words}
In this technical report, we presented our strategies to address tasks 1 and 2 of the ISIC 2019 challenge. We trained 13 state-of-the-art convolutional neural networks (CNNs) models in order to compose an ensemble of classifiers. Regarding the eight skin lesion classification, we obtained similar results to the previous competition. We may observe that in terms of balanced accuracy, the SENet architecture is the best model among the 13 trained. In this sense, the ensemble based only in the three best models performed better to our tests.

For this year, we believe the outlier detection is the most difficult point of this task. We introduced two approaches to handle the outlier class. The first one is a hierarchical classifier to detect skin images and the second one is an approach based on entropy to select any outlier. The results presented for these approaches show that the second one is able to find much more outlier samples than the first one. However, it does not means it is a better approach. In fact, we aim to improve this part in future works. For now, we are excited to see the solutions for this challenge. 

\section*{Acknowledgments}
A. G. C. Pacheco would like to thanks the financial support of the Coordena\c{c}\~{a}o de Aperfei\c{c}oamento de Pessoal de N\'{i}vel Superior - Brasil (CAPES) - Finance Code 001; T. Trappenberg acknowledges funding by NSERC.

\section*{References}
\bibliography{references}

\end{document}